\documentclass{article} 
\usepackage{iclr2015,times}
\usepackage{url}
\usepackage{graphicx}
\usepackage{epstopdf}
\usepackage{subfigure}
\graphicspath{{images/}}

\title{What Do Deep CNNs Learn About Objects? }

\author{
Xingchao Peng, Baochen Sun, Karim Ali, and
Kate Saenko\thanks{www.cs.uml.edu/\~\{xpeng,bsun,karim,saenko\}} \\
Department of Computer Science\\
University of Masschusetts Lowell\\
One University Avenue, Lowell, MA, USA \\
\texttt{\{xpeng,bsun,karim,saenko\}@cs.uml.edu}  
}

\iclrfinalcopy 

\begin{document}
\maketitle

\vspace{-0.1in}
\section{Introduction}
\label{introduction}
Deep convolutional neural networks learn extremely powerful image representations, yet most of that power is hidden in the millions of deep-layer parameters. What exactly do these parameters represent? Recent work has started to analyse CNN representations, finding that, e.g., they are invariant to some 2D transformations~\cite{fischer2014descriptor}, but are confused by particular types of image noise~\cite{nguyen2014deep}. 
%
In this work, we delve deeper and ask: how invariant are CNNs to object-class variations caused by 3D shape, pose, and photorealism?

To analyse deep representations, we treat the activations of a hidden layer as input features to a linear classifier, and test how well the classifier generalizes across intra-class variations due to the above factors.
The hypothesis is that, if the representation is invariant to a certain
factor, then similar neurons will activate whether or not that factor is present in the input image.
For example, if the network is invariant to ``cat'' texture, then it will have similar activations on cats with and without texture, i.e. it will ``hallucinate'' the right texture when given a texureless cat shape. Then the classifier will learn equally well from both sets of training data. If, on the other hand, the network is not invariant to texture, then the feature distributions will differ. As a consequence, the classifier trained on textureless cat data will perform worse. 
Because factors such as object texture and background scene are difficult to isolate using 2D image data, we rely on computer graphics to generate synthetic images from 3D object models.

\vspace{-0.1in}
\section{Exploring the Invariances of CNN features}
\label{approach}

We design a series of experiments to probe CNN invariance in the context of object detection.
For each experiment, we follow these steps: 
1) select image rendering parameters, 2) generate a batch of synthetic 2D images with those parameters, 3) sample positive and negative patches for each object class, 4) extract hidden CNN layer activations from the patches as features, 5) train a classifier for each object category, 6) test the classifiers on real images.




\noindent\textbf{CNN Model, Training, and Synthetic Data Generation.}
We adopt the detection method of~\cite{RCNN}, which uses the eight-layer ``AlexNet'' architecture with over 60 million parameters~\cite{alexnet}. 
Our hypothesis is that the network will learn different invariances, depending on how it is trained. Therefore, we evaluate two different variants of the network: one trained on the ImageNet ILSVRC 1000-way classification task, which we call {\small IMGNET}, and the same network also fine-tuned for the PASCAL detection task, which we call {\small PASC-FT}. For both networks, we extract the last hidden layer (fc7) as the feature representation. We choose to focus on the last hidden layer as it is the most high-level representation and has learned the most invariance.



We choose a subset of factors that can easily be modeled using simple computer graphics techniques, namely, \textbf{object texture} and \textbf{color}, \textbf{context/background} appearance and color, \textbf{3D pose} and \textbf{3D shape}. We study the invariance of the CNN representation to these parameters using synthetic data. 
 We also study the invariance to 3D rigid rotations using real data. 

\noindent\textbf{Object Color, Texture and Context.}
We begin by investigating various combination of object colors and textures placed against a variety of background scene colors and textures. Examples of our texture and background generation settings are shown in Table~\ref{tab:color}.

We trained a series of detectors with each of the above background and object texture configurations and tested them on the PASCAL VOC test set, reporting the average precision (AP) across categories.
The somewhat unexpected result is that the generation settings \textbf{RR-RR(28.9\%), W-RR(31.2\%), W-UG(30.1\%), RG-RR(31.2\%)} with PASC-FT all achieve comparable performance, despite the fact that \textbf{W-UG} has no texture and no context. Results with real texture but no color in the background (\textbf{RG-RR, W-RR}) are the best. This indicates that the network has learned to be invariant to the color and texture of the object and its background.




\begin{table*}[t]
\scriptsize  
\begin{center}
\begin{tabular}{|l||c|c|c|c|c|c|}
\hline
 &\textbf {RR-RR} & \textbf{W-RR}&\textbf{W-UG}  &\textbf{RR-UG}&\textbf{RG-UG}&\textbf{RG-RR}\\ 
\hline
\footnotesize{BG} & Real RGB&White & White &Real RGB&Real Gray&Real Gray\\
\hline
\footnotesize{TX} &Real RGB & Real RGB&Unif. Gray&Unif. Gray&Unif. Gray&Real RGB\\
\hline
& \includegraphics[width=0.7in]{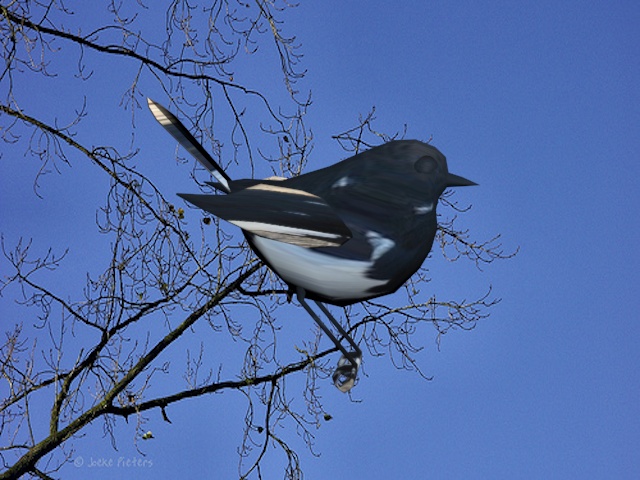}
&\includegraphics[width=0.7in]{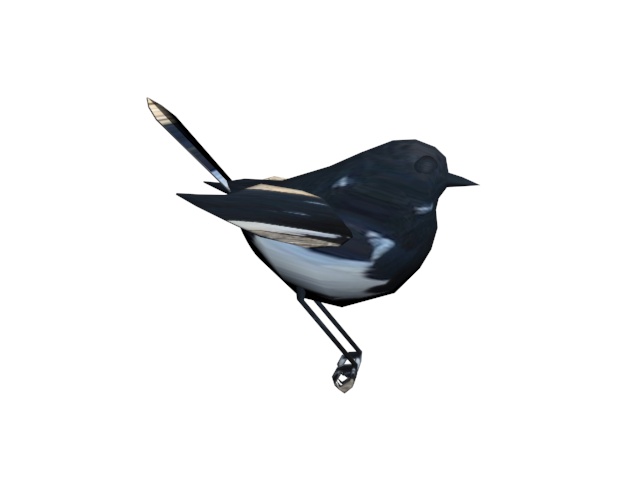}
& \includegraphics[width=0.7in]{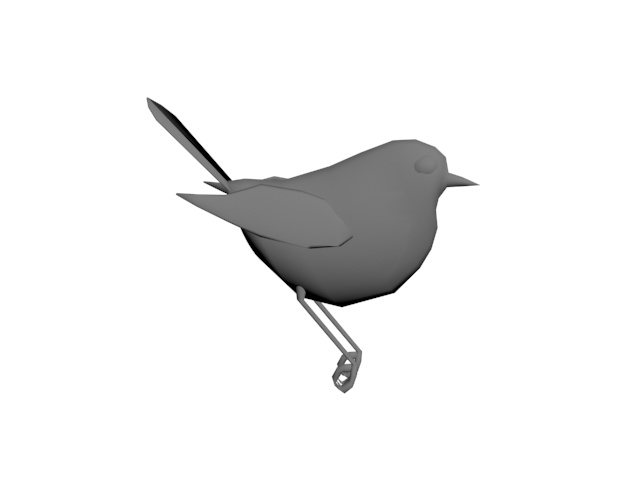}
&\includegraphics[width=0.7in]{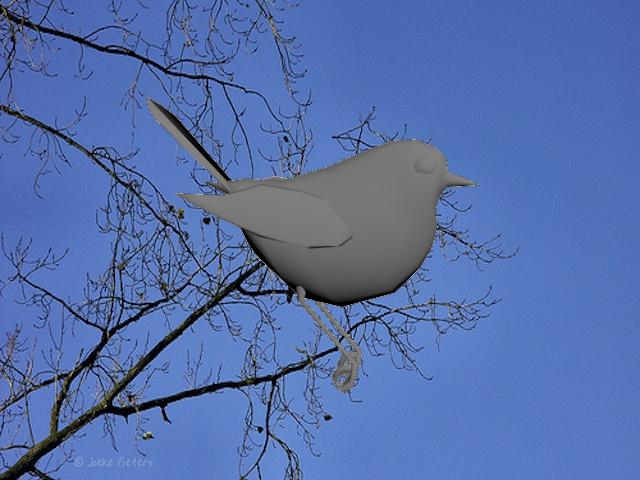}
&\includegraphics[width=0.7in]{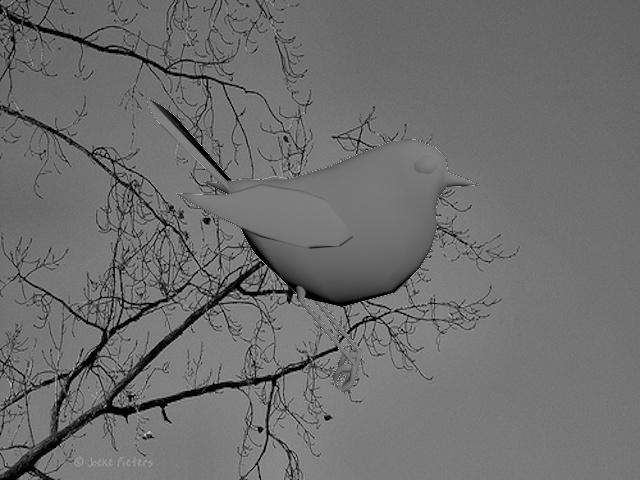}
&\includegraphics[width=0.7in]{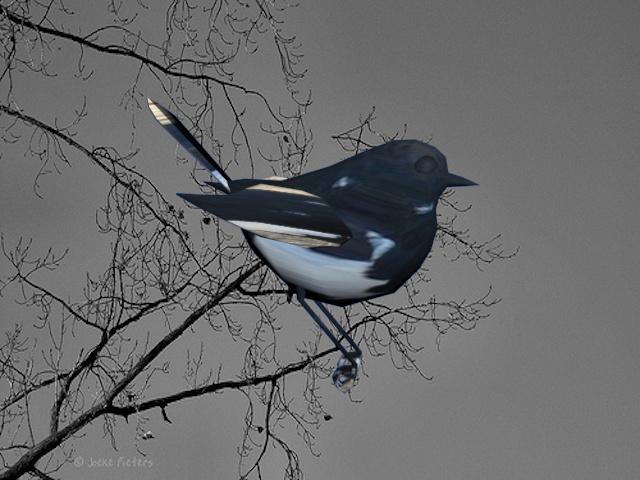}\\
\hline
\end{tabular}
\end{center}
\caption{Different configuration of background, color, texture}
\label{tab:color}
\vspace{-0.1in}
\end{table*}

\noindent\textbf{Image Pose.}
We also test view invariance on real images. We are interested here in objects whose frontal view presentation differs significantly (ex: the side-view of a horse vs a frontal view). To this end, we selected 12 categories from the PASCAL VOC training set which match this criteria. Held out categories included rotationally invariant objects such as bottles or tables. Next, we split the training data for these 12 categories to prominent side-view and front-view, as shown in Table~\ref{tab:realpose}.


We train classifiers exclusively by removing one view (say front-view) and test the resulting detector on the PASCAL VOC test set containing both side and front-views.We also compare with random view sampling. Results, shown in Table~\ref{tab:realpose}, point to important and surprising conclusions regarding the representational power of the CNN features. Note that mAP drops by less than $2\%$ when detectors exclusively trained  by removing either view are tested on the PASCAL VOC test set.



\begin{table*}[t]
\small
\begin{center}
\includegraphics[width=0.6\linewidth]{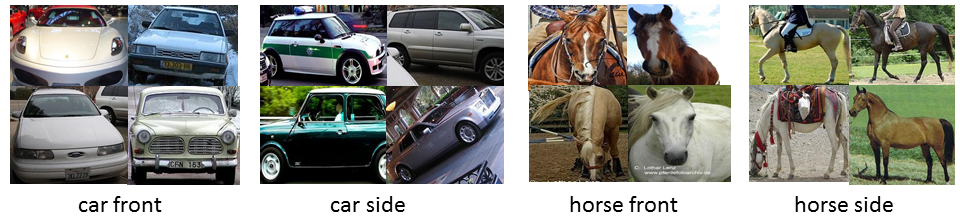} \\
\begin{tabular}
{c |c ||  p{0.25cm}  p{0.25cm}  p{0.25cm}  p{0.25cm} p{0.25cm} p{0.25cm} p{0.25cm} p{0.25cm}  p{0.25cm} p{0.25cm} p{0.25cm} p{0.40cm} |p{0.35cm}}

Net & Views &aero& bike&bird&bus&car&cow	&dog	&hrs	&mbik	&shp	&trn	&tv&mAP\\
\hline
PASC-FT & \textbf{all}&64.2&69.7&	50&	62.6&	71&	58.5&	56.1&	60.6&	66.8&	52.8&	57.9&	64.7&	61.2\\
PASC-FT &\textbf{-random}& 62.1&70.3&49.7&61.1&70.2&54.7&55.4&61.7&67.4&55.7&57.9&64.2&60.9\\
PASC-FT &\textbf{-front}&61.7&67.3&45.1&58.6&70.9&56.1&55.1&	59.0&66.1&54.2&53.3&61.6&59.1\\
PASC-FT & \textbf{-side}&62.0&70.2&48.9&61.2&70.8&57.0&53.6&59.9&65.7&53.7&58.1&64.2&60.4\\

PASC-FT(-front) & \textbf{-front}	&59.7&63.1&42.7&55.3&64.9&54.4&54.0&56.1&64.2&55.1&47.4&60.1&56.4\\
\hline
\end{tabular}


\end{center}
\caption{Results of training on different real image views. '-' represent removing a certain view.
}
\label{tab:realpose}
\vspace{-0.1in}
\end{table*}

\noindent\textbf{3D Shape.} Finally, we experiment with reducing intra-class shape variation by using fewer CAD models per category. We otherwise use the same settings as in the \textbf{RR-RR} condition with PASC-FT. 
 From our experiments, we find that the mAP decreases by about 5.5 points from 28.9\% to 23.53\% when using only a half of the 3D models. This shows a significant boost from adding more shape variation to the training data, indicating less invariance to this factor.

\section{Conclusion}

We investigated the sensitivity of convnets to various factors in the training data: 3D pose, foreground texture and color, background image and color. To simulate these factors we used synthetic data generated from 3D CAD models and a few real images.

Our results demonstrate that the popular deep convnet of \cite{alexnet} fine-tuned for detection on real images for a set of categories is indeed invariant to these factors. 



For more details and results, we refer the reader to the following paper~\cite{xpeng2015exploring}.

\footnotesize{\bibliography{iclr2015}}
\bibliographystyle{iclr2015}
\end{document}